\documentclass[journal]{IEEEtran}

\usepackage{cite}
\usepackage{amsmath,amssymb,amsfonts}

\usepackage[noend]{algorithmic}
\usepackage{algorithm}
\usepackage{graphicx}
\usepackage{textcomp}
\usepackage{subcaption}
\usepackage{textcomp}
\usepackage{xcolor}
\usepackage{mathtools}
\usepackage{makecell}
\usepackage{dsfont}
\usepackage{hyperref}
\usepackage{array}
\usepackage{bm}

\newcolumntype{P}[1]{>{\centering\arraybackslash}p{#1}}

\def\etal{\emph{~et~al. }}
\makeatother

\ifCLASSINFOpdf
\else
\fi
\begin{document}
\title{Towards Learning Generalizable Driving Policies from Restricted Latent Representations}

\author{Behrad Toghi$^{*}$, Rodolfo Valiente$^{*}$, Ramtin Pedarsani, Yaser P. Fallah
\thanks{$^{*}$Authors B. Toghi and R. Valiente contributed equally.}
\thanks{Behrad Toghi, Rodolfo Valiente, and Yaser P. Fallah are with the Department of Electrical and Computer Engineering, University of Central Florida. {\tt\small toghi@knights.ucf.edu}}
\thanks{Ramtin Pedarsani is with the Department of Electrical and Computer Engineering, UC Santa Barbara.}
\thanks{This material is based upon work partially supported by the National Science Foundation under Grant No. CNS-1932037.}}

\markboth{\normalfont{Under review in an IEEE Journal}}%
{Shell \MakeLowercase{\textit{et al.}}: Bare Demo of IEEEtran.cls for IEEE Journals}

\maketitle

\begin{abstract}
Training intelligent agents that can drive autonomously in various urban and highway scenarios has been a hot topic in the robotics society within the last decades. However, the diversity of driving environments in terms of road topology and positioning of the neighboring vehicles makes this problem very challenging. It goes without saying that although scenario-specific driving policies for autonomous driving are promising and can improve transportation safety and efficiency, they are not a universal scalable solution. Instead, we seek decision-making schemes and driving policies that can generalize to novel and unseen environments. In this work, we capitalize on the key idea that human drivers learn abstract representations of their surroundings that are fairly similar among various driving scenarios and environments. Through these representations, human drivers are able to quickly adapt to novel environments and drive in unseen conditions. Formally, through imposing an information bottleneck, we extract a latent representation that minimizes the \textit{distance} ---a quantification that we introduce to gauge the similarity among different driving configurations--- between driving scenarios. This latent space is then employed as the input to a Q-learning module to learn generalizable driving policies. Our experiments revealed that, using this latent representation can reduce the number of crashes to about half.
\end{abstract}
\begin{IEEEkeywords}
Autonomous Driving, Latent Representations, Domain Adaptation, Reinforcement Learning
\end{IEEEkeywords}
\IEEEpeerreviewmaketitle

%
%
\section{Introduction}
\label{sec:intro}
\IEEEPARstart{A}{utonomous} vehicles promise major improvements in the safety and efficiency of transportation. In this work, we focus on the decision-making and behavior planning aspects of autonomous driving and address an important issue that can bring urban autonomous driving significantly closer to reality. Urban autonomous driving is particularly challenging due to the high degree of stochasticity and diversity in the configurations that vehicles are placed as well as in road topologies. Complex and competitive scenarios such as passing intersections, navigating through roundabouts, and exiting/merging in highways require driving policies that not only ensure safety, but also can produce socially-desirable outcomes~\cite{toghi2021social}. During the last decades, a plethora of solutions have been presented in the literature that address each of the above-mentioned scenarios separately and have shown decent performance in handling them in isolation~\cite{toghi2018multiple,toghi2019analysis,toghi2019spatio,toghi2020maneuver,8690570, saifuddin2020performance,mahjoub2019v2x, shah2019real, mahjoub2018driver, valiente2019controlling, valiente2020connected}. However, a realistic and scalable solution does not comprise of ad-hoc solutions for each scenario but rather requires a generalizable driving policy\cite{toghi2021cooperative, toghi2021altruistic}.

As human drivers, we do not learn ad-hoc behaviors for each driving scenario and environment configuration that we face. Human drivers are often able to handle novel environments and unseen road topologies by leveraging their prior knowledge and experience. This fact inspires us to hypothesize the existence of general representations (and therefore learnable policies) that incorporate high-level abstract information and are independent of particular driving scenarios. In other words, we argue that learning general and scenario-independent driving behaviors is possible if and only if there exists a space in which different driving scenarios, e.g., crossing an intersection, exiting/merging to a highway, entering a roundabout, are represented similarly. Such representation space removes the dependency on the driving scenario and enables a human or an AI agent to learn behaviors that can expand to novel and unseen environments. Intuitively, a human driver perhaps thinks in terms of \textit{free space}, \textit{time headway}, or \textit{heading} rather than meticulously analyzing each unique driving scenario. Therefore, our first key insight is as follows,
\begin{itemize}
    \item[] \textit{Transforming the driving scene into a scenario-independent latent space representation can enable multi-task learning for autonomous driving agents as they receive only the information that is essential for the task of driving and hence become agnostic to the specific driving scenario.}
\end{itemize}

A major question that emerges after the previously introduced concepts and solutions is how these methods can be generalized to novel and unseen environments. Although the existing works on autonomous vehicle navigation provide valuable methods and algorithms to address complex driving scenarios and enable the agent to learn safe and efficient driving policies in a-hoc and limited scenarios, it is extremely important to investigate the sensitivity of these policies with regards to change in road topology and mission. In this work, we explore the possibility of employing a learned latent space as the state representation for training reinforcement learning (RL) agents, and the proposed solution showed improvement in the generalization capabilities of the resulting policies.

We list the contributions of this paper as follows:
\begin{itemize}
  \item A latent representation is learned through imposing an information bottleneck in an encoder-decoder structure that encompasses the most crucial information that is needed for the policy learning module;
  \item In addition to the information bottleneck, we provide a similarity metric between driving scenarios and impose a regularization term that is intended to minimize this metric in the learned latent space;
  \item Our experiments reveal that a general latent space exists that can be used as the input to the Q-learning module and enable agents to generalize to unseen and novel environments.  Using our solution, the number of failed scenarios was reduced by about 50\%.
\end{itemize}

%
%
\section{Related Work}
\label{sec:relatedworks}
\subsection{Urban Autonomous Driving Solutions}
A large spectrum of research work has been done on urban autonomous driving, particularly to overcome complex and stochastic scenarios such as roundabout, intersection and highway merging ramps. Azimi\etal has done extensive work on intersection management and deriving reliable protocols for connected automated vehicles to pass intersections, including their Ballroom Intersection Protocol (BRIP) method~\cite{azimi2013reliable, azimi2015ballroom}. Wang\etal and Dong\etal similarly focus their work on another competitive urban driving scenario, namely highway merging ramp~\cite{wang2017formulation, dong2017intention}. In their recent work, Hoermann\etal overcome challenging "downtown scenarios" that include pedestrians, bikes, and motor vehicles, interacting with each other through a combination of Bayesian filtering techniques for environment ~\cite{hoermann2018dynamic}. Chen\etal argue that the existing manually designed driving policies are not capable of handling complex and highly-stochastic urban scenarios. Instead, they rely on a data-driven deep reinforcement learning solution and evaluate their method in a dense roundabout. Their high-definition simulation episodes have demonstrated the superior performance of RL-based solutions~\cite{chen2019model}.

Chen\etal further extend their insights and explore the possibility of end-to-end learning. They jointly train a sequential environment model in conjunction with the reinforcement learning process to extract driving policies for urban scenarios~\cite{chen2021interpretable}. A major challenge in urban autonomous driving is the fact that autonomous agents might behave over-defensive if they sense potential dangers and thus lose their edge in competitive driving scenarios. Zhan\etal propose a non-conservatively defensive maneuver planning scheme that comprises of two stochastic and deterministic sub-modules and shown enhanced performance in handling urban environments~\cite{zhan2016non}. Huang\etal combine the well-known artificial potential fields (APF) method with a mesh-grid resistance network and present a novel motion planning and tracking solution for autonomous driving in urban scenes~\cite{huang2019motion}. Chen\etal rely on Iterative linear quadratic regulator (ILQR) method and formulate the urban autonomous driving as a constrained ILQR (CILQR)~\cite{chen2017constrained, chen2019autonomous}. 

\subsection{Representation Learning}
In their recent work, Tschannen\etal state that unsupervised learning of useful representations is a major challenge in AI and present an in-depth survey of recent advances in representation learning with a focus on autoencoder-based models~\cite{tschannen2018recent}. Plebe~\etal tackle the problem of visual prediction in driving scenarios and present a manipulable latent representation that enables their solution to generalize in different urban driving scenarios~\cite{plebe2020road}. Their work is inspired by the theory that the human brain utilizes ensembles of neurons for information compression and finding abstract concepts from visual experience and coding them into compact representations. Li\etal propose a method that captures non-linear relationships among complex features of a network by exploiting a deep unsupervised generation algorithm, namely variational autoencoder (VAE)~\cite{li2017variation}. Chen\etal introduce a latent space that is learned via sequential latent representation learning and demonstrate that an end-to-end perception module that utilizes this latent space can handle various tasks such as tracking, detection, and localization at the same time while requiring minimum human engineering~\cite{chen2020end}.

Kargar\etal ~\cite{kargar2020efficient} suggest using low dimensional and rich feature representations of an agent's observations by training an encoder-decoder architecture that is capable of predicting multiple application relevant factors such as trajectories of other agents. Authors reveal that such multi-head architecture generates more efficient and more informative representations. Ma\etal demonstrate that a latent state that is inferred explicitly, is capable of encoding spatio-temporal relations. They employ this latent space to study the interplay between minor cues in driving environments that can lead to major differences in the outcome~\cite{ma2021reinforcement}. On the other hand, Xi\etal, Losey\etal, and Karamcheti\etal take a different approach and apply latent representation learning in applications such as partner modeling and human-robot interaction \cite{xie2020learning, losey2021learning, karamcheti2021lila, karamcheti2021learning}.

%
%
\section{Preliminaries}
\label{sec:preliminaries}
In this section, we introduce our formal notation and provide a brief overview of the background knowledge that is imperative to grasp the concepts discussed in the rest of the paper.
\subsection{Partially Observable Markov Decision Processes}
The decision-making process of an intelligent agent within a stochastic environment can be formally described as a Markov decision process (MDP) defined by the tuple $\mathcal{M} \coloneqq (\mathcal{S}, \mathcal{A}, T, R, \gamma)$. At a given time, the agent's configuration is described by its state $s \in \mathcal{S}$, it takes an action within the action space $a \in \mathcal{A}$ and transits to a new state $s'$ based on the state transition probability $T(s'|s, a)$ and receives a reward $R(s, a)$. It is common to assume that the dynamics of the environment hold a $\text{1}^{\text{st}}$-order Markov property, i.e.,
\begin{equation}
\label{equ:markov_property}
\Pr\{{s}_{t+1}=s'|{s}_t, a_t\} = \Pr\{{s}_{t+1}=s'|{s}_t, ..., {s}_0, a_t\}
\end{equation}

The probability distribution over actions in $\mathcal{A}$ at a given state is known as a policy $\pi (s)$. The goal is to derive a distribution that maximizes the discounted sum of future rewards over an infinite time horizon, i.e., an optimal policy $\pi^*: \mathcal{S} \to \mathcal{A}$. This optimal policy always exists for an environment with stationary transition probabilities and rewards,
\begin{equation}
\label{equ:optimalpolicy}
\pi^* \coloneqq \underset{\pi \in \theta}{\arg\max} \; \mathbb{E} \left[ \sum_{i=0}^{\infty} \gamma^i r(s_i,\pi(s_i)) \right]
\end{equation}
in which, $\gamma \in [0,1)$ is a discount factor and $\theta$ describes the set of all admissible Markov chains that can be extracted from $\mathcal{M}$. The optimal policy maximizes the action-value function, i.e., $\pi^*(s) = \arg\max_a Q^* (s,a)$, where
\begin{equation}
\label{equ:qfunction}
Q^\pi(s,a) \coloneqq \mathbb{E} [\sum_{i=1}^\infty \gamma^i r(s_i, \pi (s_i)) |s_0=s, a_0=a]
\end{equation}
and the optimal action-value function can then be derived using the Bellman equation,
\begin{equation}
\label{equ:bellmanequ}
Q^*(S,a) = \mathbb{E}_{s'\sim P(.|s,a)} \left[ r(s,a) + \max_a \gamma Q^*(s',a') \right]
\end{equation}
\subsection{Solving POMDPs with Unknown Dynamics}
If the MDP is fully known, dynamic programming algorithms such as value and policy iteration can be used to recursively solve for the optimal action-value function $Q^*$. However, in real-world problems, the dynamics of the environment and reward function are usually not known and an agent only has access to a local observation that is correlated to the underlying state, i.e., a partially-observable Markov decision process~(POMDP). Reinforcement Learning (RL) provides a possibility to solve POMDPs with unknown reward and state transition functions through continuous interaction with the environment.

Mathematically, RL algorithms such as temporal difference (TD) learning enable agents to update the value function from such interactions with the environment without explicitly knowing the full MDP, as follows
%
\begin{multline}
\label{equ:TDlearning}
Q_{i+1}(s,a) - Q_i(s,a) =\\
\alpha_i \left[ r(s, \pi (s)) + \gamma \max_{a'} Q_i(s',a') - Q_i(s,a) \right]
\end{multline}
%
in which $\alpha_i$ is learning rate at the $i$th iteration \cite{lee2020optimization}.

\subsection{Deep Q-networks}
\label{sec:dqn_preliminaries}
Parameterizing the action-value function using a function approximator, i.e., $\Tilde{Q}(.) \cong Q(.;\textbf{w})$, makes learning more generalizable policies and scaling to larger state-spaces possible. Parameters $\textbf{w}$ can be learned through mini-batch gradient descent steps,
\begin{equation}
\label{equ:Qlearning}
\textbf{w}_{i+1} = \textbf{w}_i + \alpha_i \hat{\nabla}_\textbf{w} \mathcal{L}(\textbf{w}_i)\\
\end{equation}
where, the $\hat{\nabla}_\textbf{w}$ operator estimates the gradient at $\textbf{w}_i$.

Deep neural networks are widely considered as function approximators and are also applicable to the Q-learning algorithm. Mnih\etal. leveraged this idea and introduced Deep Q-networks, a solution suitable for RL problems with large state-spaces~\cite{mnih2013playing}. DQN builds up on two major ideas, namely employing an \emph{experience replay buffer} to generate training samples, and using two separate networks during training. The key idea is to stabilize the training process by updating the greedy network at each training iteration to compute the optimal Q-value and using another less-frequently updated target network. The loss function in Equation~\ref{equ:Qlearning} can be written as
\begin{equation}
\label{equ:GD_loss_target}
\mathcal{L}(\textbf{w}_i) = \mathbb{E}[r+\gamma \underset{a}{\max} Q^*(s',a';\textbf{w}^\circ) - Q^*(s,a;\textbf{w})]^2
\end{equation}
where $\textbf{w}^\circ$ is the target network which periodically gets updated during the training.

\subsection{Denoising autoencoders}
\label{sec:denoising_ae}
Vincent\etal proposed the denoising autoencoders which are auto-associator networks that are trained to learn hidden representations that are sufficient for re-creating the input. The key insight is that a random noise is added to the input and the network is forced to learn to not only reconstruct the input, but also denoise the corrupted input and extract the original input~\cite{vincent2008extracting}. This is particularly useful for our application, as we deal with real-time observations that are most likely generated by sensor fusion over camera, LiDAR and radar inputs and therefore can be very noisy. Formally, we denote the true observation of the environment as $\boldsymbol{x}$, the noisy observation $\tilde{\boldsymbol{x}}$ can then be defined through a stochastic process $q_D$ as follows,
\begin{equation}
\label{equ:ae_stochastic_process}
\tilde{\boldsymbol{x}} \sim q_D(\tilde{\boldsymbol{x}}|\boldsymbol{x})
\end{equation}

Using a matrix of weights $\boldsymbol{W}$ and bias vector $\boldsymbol{b}$ and applying a non-linearity $\sigma_y$, the output of the encoder creates a latent representation $\boldsymbol{y}$ which can be computed as,
\begin{equation}
\label{equ:ae_encoder}
\boldsymbol{y} = \sigma_y (\boldsymbol{W}\tilde{\boldsymbol{x}}+\boldsymbol{b})
\end{equation}
We assume the encoder and decoder networks share the same set of weights. Thus, the output of the decode, i.e., the reconstruction $\boldsymbol{z}$ can be re-created through applying the transpose of the weight matrix as follows,
\begin{equation}
\label{equ:ae_decoder}
\boldsymbol{z} = \sigma_z (\boldsymbol{W}^{\boldsymbol{T}}\tilde{\boldsymbol{y}}+\boldsymbol{c})
\end{equation}
in which, $\boldsymbol{c}$ is the visible bias vector and $\sigma_z$ is the non-linearity. Such encoder-decoder structure can be trained to reconstruct the original input $\boldsymbol{x}$ from the noisy input. Using the following loss function, a standard back-propagation routine can be used to learn the encoder-decoder weights,
\begin{equation}
\label{equ:ae_loss}
L_2(\boldsymbol{x}, \boldsymbol{z}) = \sum_i (x_i-z_i)^2
\end{equation}

As proposed by Gehring\etal, we train 2 autoencoders sequentially~\cite{gehring2013extracting}. The second autoencoder is trained to learn the hidden representation of the first autoencoder. The second autoencoder essentially learns a probability distribution over the weights of the first autoencoder. A cross-entropy loss can be used for training the second autoencoder,
\begin{equation}
\label{equ:ae_cross_entropy_loss}
L_H(\boldsymbol{x}, \boldsymbol{z}) = \sum_i x_i \log z_i + (1-x_i)\log z_i
\end{equation}
Authors suggest training a stack of autoencoders in the proposed fashion and using the top autoencoder network as the bottleneck.

\begin{figure}[htbp]
    \centering
    \includegraphics[width=0.47\textwidth]{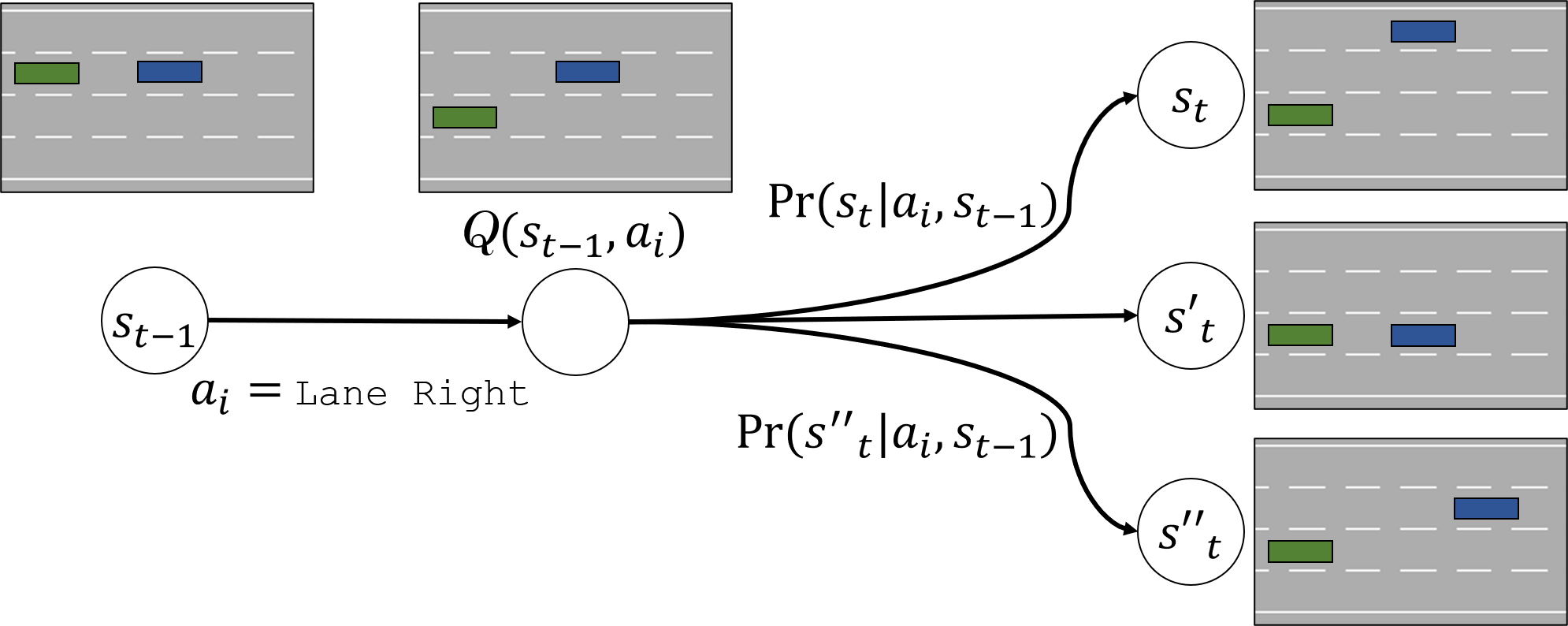}
    \caption{The Q-state}
    \label{fig:q_state}
\end{figure}
%

%
%
\section{Problem Statement}
\label{sec:problem_statement}
As elaborated in Section~\ref{sec:intro}, our goal in this work is to derive a driving policy that is not limited to a particular road topology or configuration of the vehicles. Such scenario-independent driving policy is expected to be able to generalize and operate in unseen and novel environments. We argue that drivers are well able to achieve this task through thinking in an abstract space that can generalize to any given scenario, e.g., looking at free spaces, general features such as their distance to the middle lane, and most importantly, traffic rules. Inspired by this argument, we explore the possibility of finding a latent space which encompasses abstract representations of driving scenes. Our key insight is that the reason that a driving policy trained in scenario A, e.g., an intersection, does not perform well in scenario B, e.g., highway merging ramp, is the difference between road topologies, possible configurations of agents, and possible trajectories. 
\subsection{Learning a Similarity-maximizing Latent Representation}
Existing works employ engineered inputs such as heatmaps, or multi-channel representations similar to what the authors have used in~\cite{toghi2021cooperative} to feed the ego vehicle's observations into the Q-learning module. We introduce a similarity metric that measures the distance between different scenarios and seek a latent space that maximizes this similarity among different driving scenarios. This latent space is then used to represent the ego vehicle's observation of the environment. We also aim to consolidate planning and prediction via using this latent representation as the input to the reinforcement learning module to eventually learn driving policies. In contrast with the engineered state representation approach in the literature, in this work, we rely on data to learn the most efficient and useful information that can be used at the Q-learning stage. This data-driven representation is then expected to minimize the similarity distance between different driving scenarios. Particularly we choose a set of 5 different road topologies, namely roundabout, intersection, highway merging, highway exiting, and highway cruising. Each scenario is then further randomized in terms of vehicles' speed and position to gauge our method's capability in generalization.

\begin{figure*}[t]
  \centering
  \includegraphics[width=.99\textwidth]{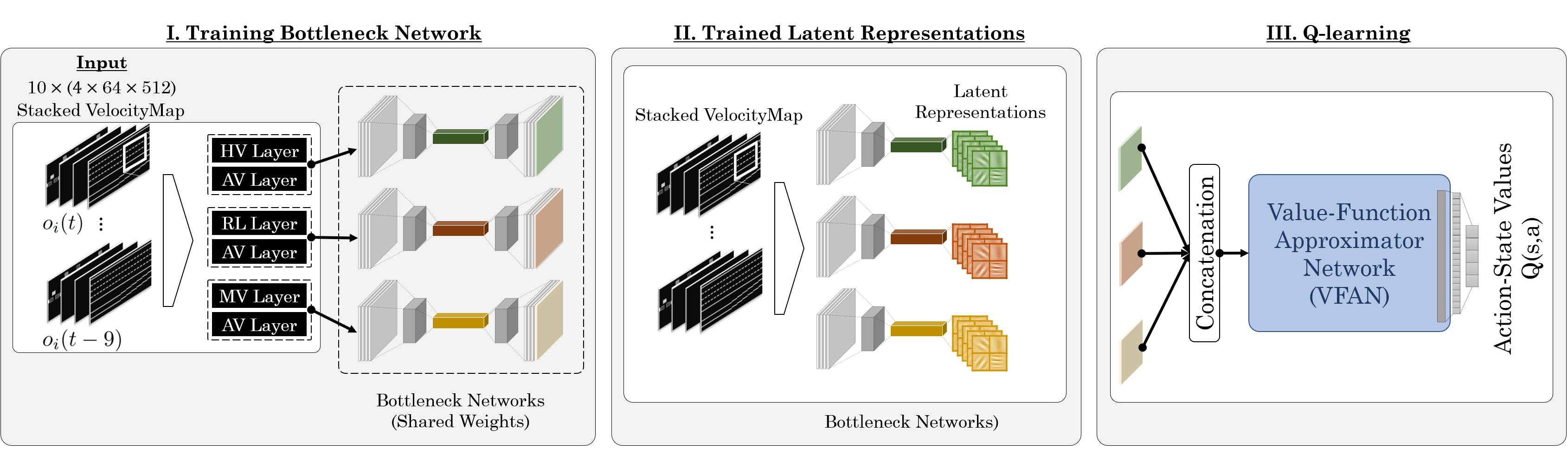}
\caption{{\textit{(HV: Human-driven Vehicles, AV: Autonomous Vehicles, MV: Mission Vehicle, RL: Road Layout)} The Bottleneck Network (BNN) ensures the most crucial information pass the data pipeline and aims to maximize the similarity across different driving scenarios. The learned latent representation is used for learning driving policies via Q-learning.}}
  \label{fig:overall_system_diagram}
\end{figure*}
\subsection{Joint Planning \& Prediction with Reinforcement Learning}
Following the Markov Decision Process notation that we introduced in Section~\ref{sec:preliminaries}, all vehicles take an action at a time step $t$ which evolves the state of the environment from an initial state to a goal state. Despite the stochasticity of this transition, the underlying probability distribution defines the dynamics of the environment, including the behavior of humans and autonomous agents. This probability distribution $\Pr(s_t=s'|s_{t-1}=s,a_{t-1}=a)$ depends on the actions of all vehicles as well as the dynamics of the world itself. To further elaborate, consider the simple example from Figure~\ref{fig:q_state} where an autonomous agent takes an arbitrary action $a_i$ at the state $s_{t-1}=s^0$ and is therefore at the intermediate Q-state $Q(s^0, a_i)$. The next state of the environment $s_t$ depends on both the actions taken by AVs and the actions of HVs, i.e., the policy of AVs and the model describing the behavior of a human driver. Hence, including humans in the scenario adds significant complexity to the MDP as their behavior is often hard to model and can also vary with time. Additionally, the existing behavior models of human drivers are mostly derived in the absence of AVs and whereas the prior work in the literature reveals that humans act differently when AVs are around them. The modular approach with independent modeling (prediction) and planning (decision-making) components is therefore not ideal in the presence of humans, which motivates us to combine the two in our approach.

A core part of decision-making for autonomous vehicles in the presence of humans is understanding and anticipating the behavior of human drivers. We build on this virtue and seek a latent space that not only can encompass general driving representations but also embeds the human behavior modeling and inter-agent relations, as explored by Toghi\etal~\cite{toghi2021social} and Schwarting\etal~\cite{schwarting2019social}. We believe that the behaviors of human drivers as well as the inter-agent relations can also be explained in a general and scenario-independent manner, similar to our previous discussion. Our second key insight expands our work to a joint prediction and planning regimen. Particularly, we hypothesize that a generalizable latent space can incorporate inter-agent relations and enable the ego vehicle to predict their behaviors in a fashion that does not depend on the driving scenario.
%
%

\section{Proposed Solution}
\label{sec:solution}

Our proposed solution comprises of two sub-systems, namely an information bottleneck that is trained to extract similarity-maximizing latent representations of driving scenes as well as a Q-learning module that takes the latent representation as its input and generates a probability distribution over possible actions, i.e., a driving policy. This proposition is built on two main pillars, inspired by the work from Gupta\etal~\cite{gupta2017cognitive}. First, we believe the most efficient and useful state representation is not hand-engineered but rather learned from data. Second, we hypothesize that consolidating planning and prediction modules for decision-making in an autonomous vehicle can improve their generalization ability in handling novel and unseen topologies and configurations. 

Our architecture includes a bottleneck encoder-decoder structure as well as a 3-dimensional convolutional neural network (CNN) that serves as a function approximator to estimate the action-value function defined in Eq~\eqref{equ:qfunction}. Figure~\ref{fig:overall_system_diagram} illustrates an overview of our solution. The input to this system is a noisy spatio-temporal state representation, i.e., Stacked VelocityMaps as defined in Section~\ref{sec:state_action_space}, and the output is a probability distribution over actions $a \in \mathcal{A}$ at the given state $s$.

\begin{figure}[b]
  \centering
  \includegraphics[width=.49\textwidth]{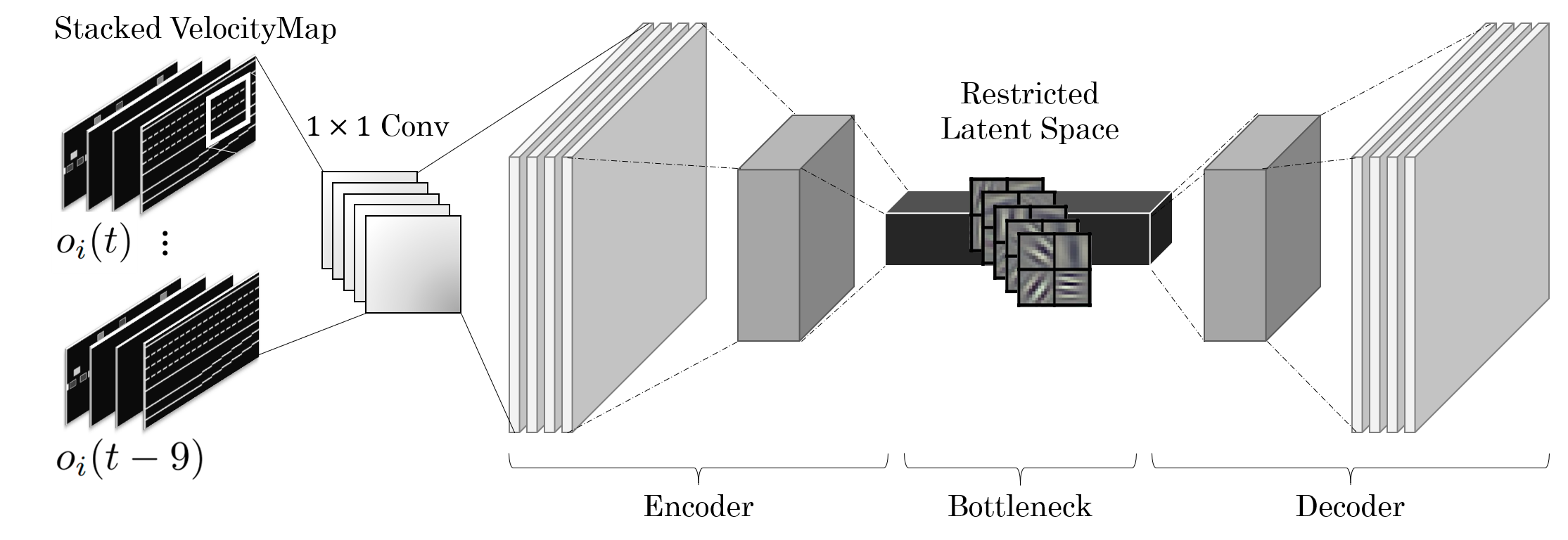}
\caption{{The Bottleneck Network (BNN) ensures only the most crucial information pass the data pipeline and removes the scenario dependency by maximizing the similarity across driving scenes from different scenarios.}}
  \label{fig:AEN}
\end{figure}
\subsection{State-space and Action-space}
\label{sec:state_action_space}
It is a common practice to study robot navigation problems from different abstraction levels ranging from low-level continuous control signals to trajectory planning and eventually behavioral planning. In this work, we view the problem in terms of high-level meta-actions. Formally, we define a discrete action space $\mathcal{A}$ where the $i$th agent's action can be $a_i \in \mathcal{A}_i = [\texttt{Left}$, $\texttt{Idle}$, $\texttt{Right}$, $\texttt{Accelerate}$, $\texttt{Decelerate}]^T$. Our driving simulator, introduced in Section~\ref{sec:drivingsimulator}, renders these discrete actions into smooth and plausible trajectories and utilizes a PID controller to generate low-level steering and throttle signals that enable a car to follow the desired trajectory. 

We introduce an intuitive and rich state representation that carries spatio-temporal information about the scene and naturally is contaminated with sensory noise. The Stacked \textit{Multi-channel VelocityMap} shown in Figure~\ref{fig:AEN} shows the position of autonomous vehicles (AV layer) and human-driven vehicles (HV layer) with their relative Frenet longitudinal velocity embedded in the pixel values. To better control the dynamic range of the pixel values in VelocityMaps, we employ a clipped logarithmic function which demonstrated enhanced performance compared to the linear mapping. The mapping from the relative speed to pixel values is
\begin{equation} \label{equ:logmapping}
Z_j = 1 - \beta \log (\alpha |v_j^{(l)}|) \mathds{1}(|v_j^{(l)}|-v_0) 
\end{equation}
in which $Z_j$ denotes the pixel value of the $j$th vehicle/object in the state, $v^{(l)}$ is its relative Frenet longitudinal speed from the $k$th vehicle's point-of-view, i.e., $\dot{l_j}-\dot{l_k}$, $v_0$ is speed threshold, $\alpha$ and $\beta$ are dimensionless coefficients, and $\mathds{1}(.)$ is the Heaviside step function. The rationale behind this design choice is to give more importance to neighboring vehicles with smaller $|v^{(l)}|$ and almost disregards the ones that are moving either much faster or much slower than the ego. Additional channels in VelocityMaps also embed the road layout (RL layer) as well as the position and the absolute speed of the ego vehicle (MV layer).

\begin{figure}[b]
  \centering

  \includegraphics[width=0.46\textwidth]{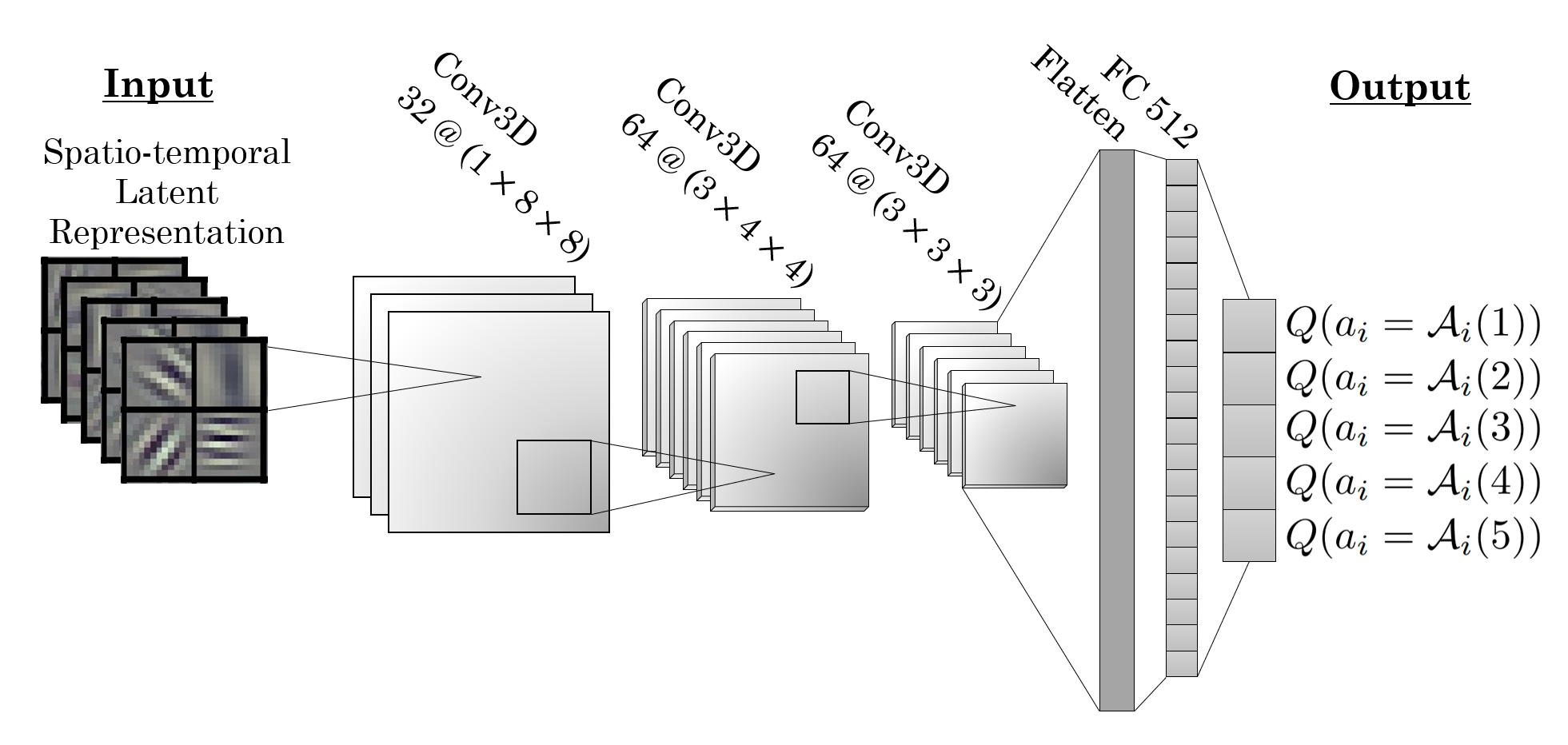}
  \caption{{3D Convolutional Value Function Approximator Network (VFAN) takes in the learned latent representations and outputs a probability distribution over actions.}}
  \label{fig:VFAN}
\end{figure}
\subsection{Information Bottleneck: Restricted Latent Space}
Starting with the autoencoder bottleneck structure, as demonstrated in Figure~\ref{fig:AEN}, a 3D Convolutional architecture is trained with mixed batches of observations 
 
\begin{figure}[b]
  \centering

  \includegraphics[width=0.4\textwidth]{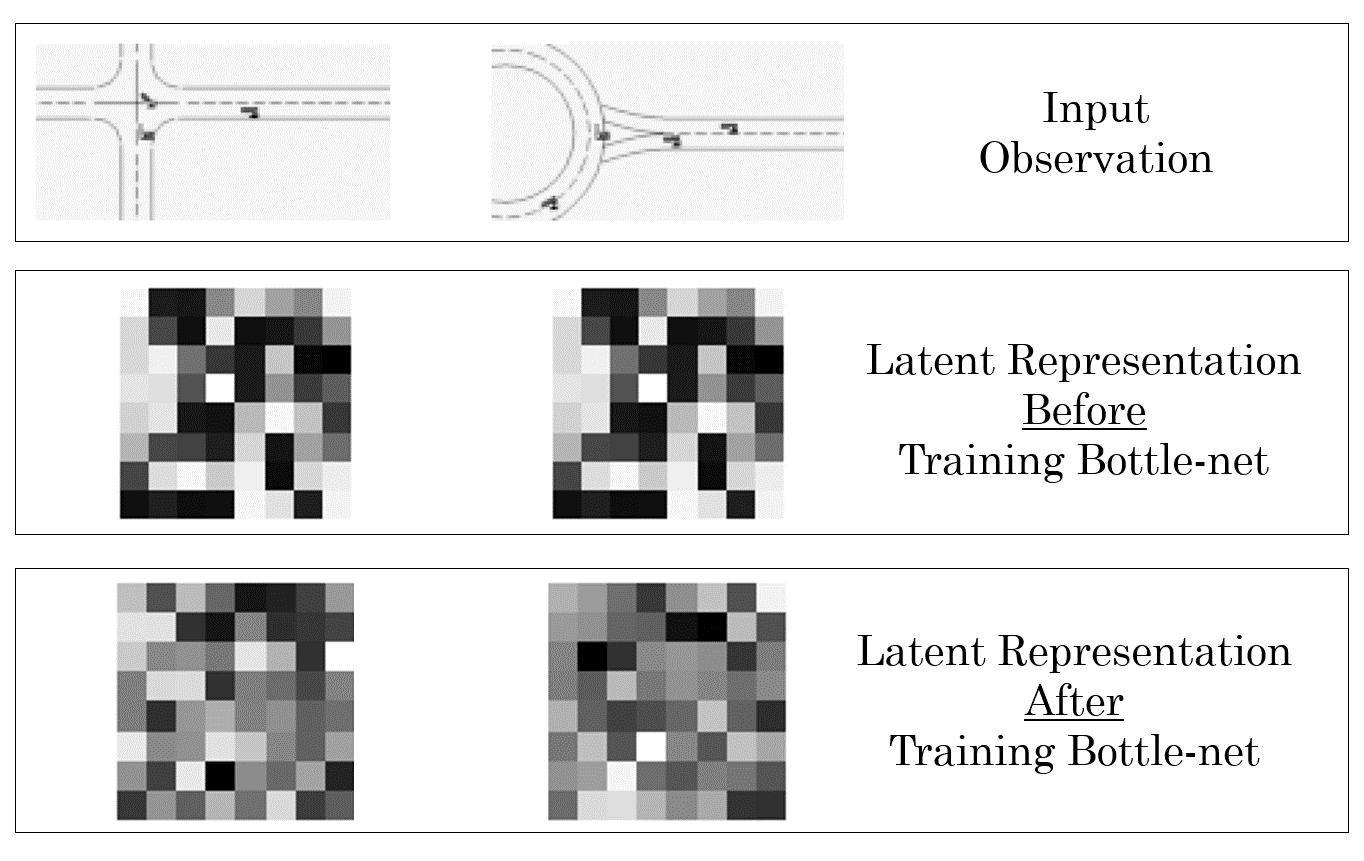}
  \caption{{Latent Representation learned through an information bottlneck are forced to grasp the most crucial information that describes the scene.}}
  \label{fig:latent_rep}
\end{figure}

Reiterating on our key insights, we seek a latent representation that encompasses the scenario-independent information required for multi-agent decision-making. We add an intermediate module to the typical deep reinforcement learning architecture to remove the dependency from driving scenarios and, in effect generalize it through a dedicated generalizing latent representation. For this purpose, we study an approach that relies on an information bottleneck (Figure~\ref{fig:AEN}) on the pipeline from our feature extraction to value-function approximation (Figure~\ref{fig:VFAN}). Imposing such bottleneck on the data flow and training encoder-decoder using the loss function defined in Section~\ref{sec:preliminaries} incentivizes the networks to only pass the most important information that explains the driving scene, positioning of the vehicles, as well as inter-agent coordination and interactions to the value-function network. We employ an encoder-decoder architecture for the bottleneck that is pre-trained as an autoencoder over mixed instances of state-space representation from the 5 driving scenarios we introduced before, i.e., roundabout, intersection, highway merging, highway exiting, and highway cruising. The pre-trained weights are then used as a starting point for training the bottleneck network over shuffled batches of different driving scenarios. This mixed training regimen ensures that the data bottleneck only passes the information that is common among driving scenarios and essential for the value-function network. Figure~\ref{fig:latent_rep} demonstrates an example of the latent representation showing two contrasting roundabout and intersection topologies.

Two major questions come up in this part, which we will investigate in Section~\ref{sec:results}. First, is it possible to stabilize the training with mixed examples from different driving scenarios and what is the guarantee for convergence? It is unclear how feeding our training pipeline with shuffled inputs from different driving scenarios can lead to a unified latent representation that is agnostic to driving scenarios. In order to enforce the bottleneck network to be independent of the driving scenario, we feed it with batches containing random instances of driving scenarios and thus the network is forced to extract inter-agent relations that are common among the samples of each batch, which are in fact from different driving scenarios. The second question is whether our desired general latent space representation exists? To extract latent spaces that capture the most useful information for driving in unseen environments as well as the inter-agent relations, we separately feed the bottleneck network (BNN) with pairs of observation components, as shown in Figure~\ref{fig:overall_system_diagram}. Specifically, we extract spatio-temporal embeddings from 3 pairs of input layers: 1) HV+AV 2) RL+AV 3) MV+AV, where, HV: human-driven vehicle VelocityMap layer, AV: autonomous vehicle VelocityMap layer, RL: road layout layer, MV: mission vehicle layer. These embeddings are then passed through the bottleneck networks with shared weights for purification and their concatenated output is fed to the value-function approximator network (VFAN).

%
%
\section{Experiments and Results}
\label{sec:results}

\begin{figure}[t]
  \centering

  \includegraphics[width=0.4\textwidth]{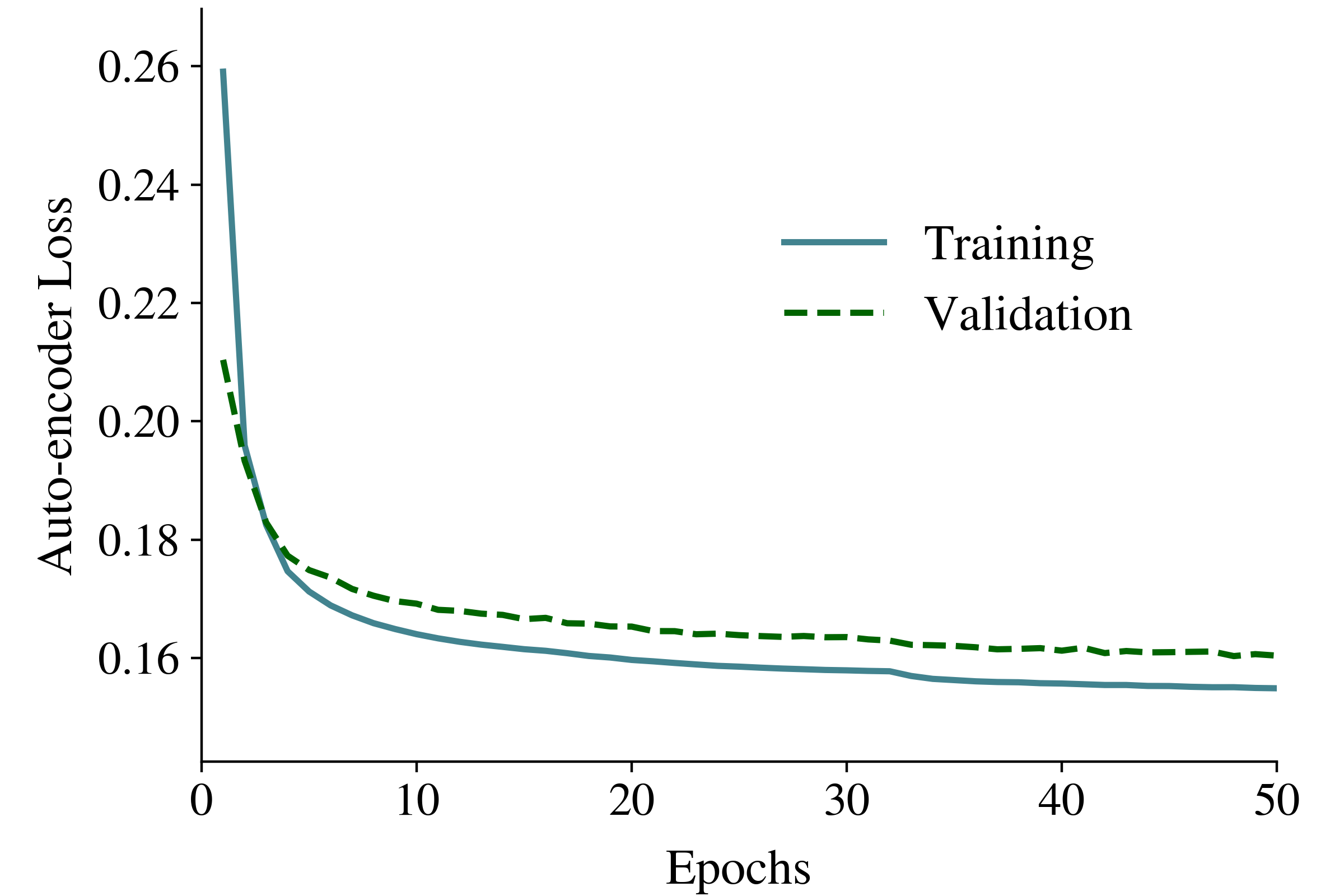}
  \caption{{The autoencoder is trained for 100 epochs over a shuffled dataset of driving instances from roundabouts, intersections, highway exiting, and highway merging. The change in loss is not visually noticeable after epoch 50.}}
  \label{fig:ae_loss}
\end{figure}
\begin{figure}[t]
  \centering

  \includegraphics[width=0.45\textwidth]{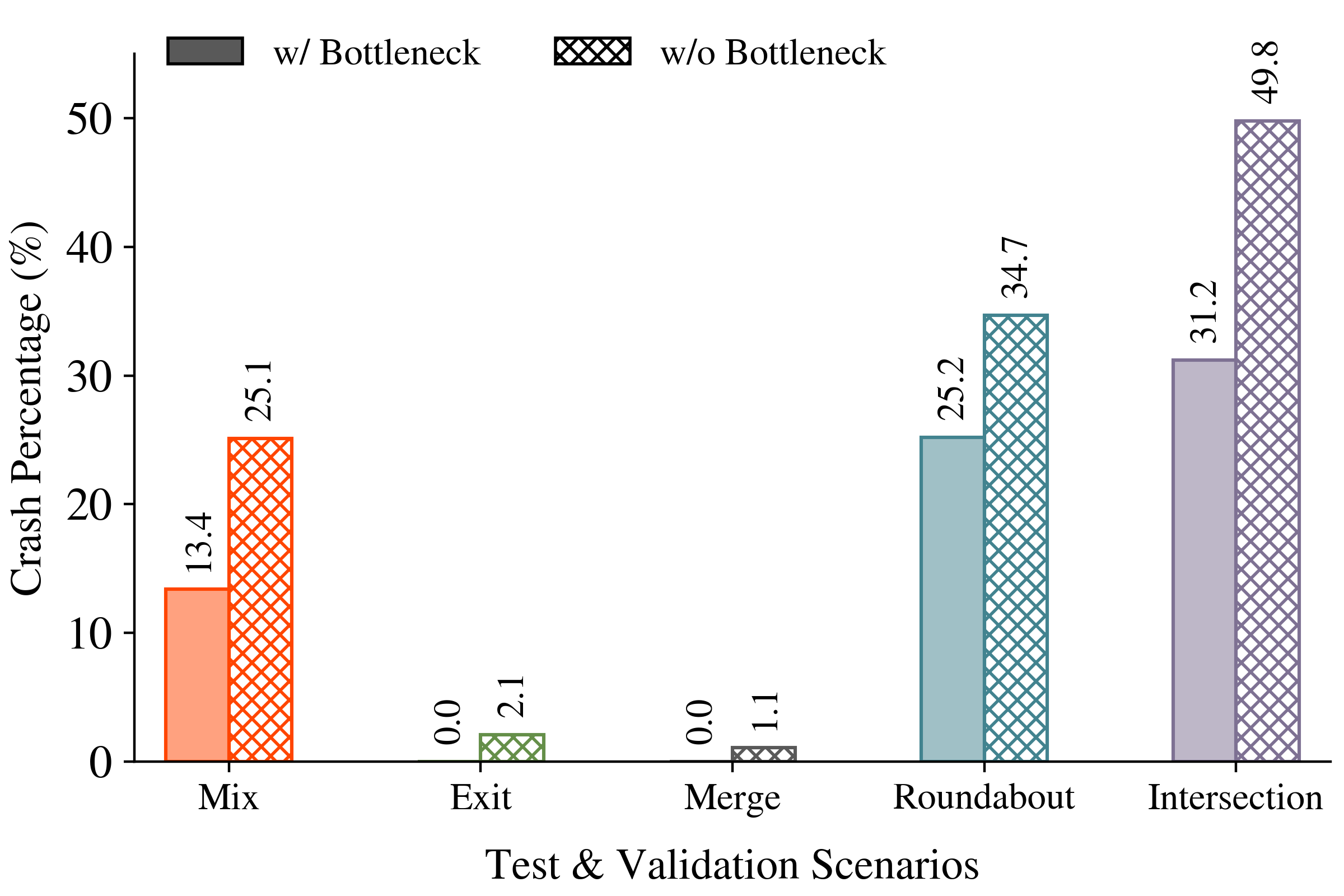}
  \caption{{Performance enhancement resulted from using the bottleneck network.}}
  \label{fig:effect_bottleneck}
\end{figure}

\subsection{Computational Details}
With our Python implementation using PyTorch and OpenAI Gym, each training episode takes about 0.3 seconds on a cluster node with an NVIDIA Tesla V100 GPU and a Xeon 6126 CPU @ 2.60GHz. The training process has been repeated several times to ensure all runs converge to similar emerging behaviors and policies. Training the bottleneck network for 100 epochs and the Q-learning network Conv3D network for 5,000 episodes took approximately 10 hours with our experiment setup. Overall, we spent $\sim$2,000 GPU-hours to train the autoencoder, Q-learning network, and tune the hyperparameters.

\begin{figure}[b]
  \centering

  \includegraphics[width=0.45\textwidth]{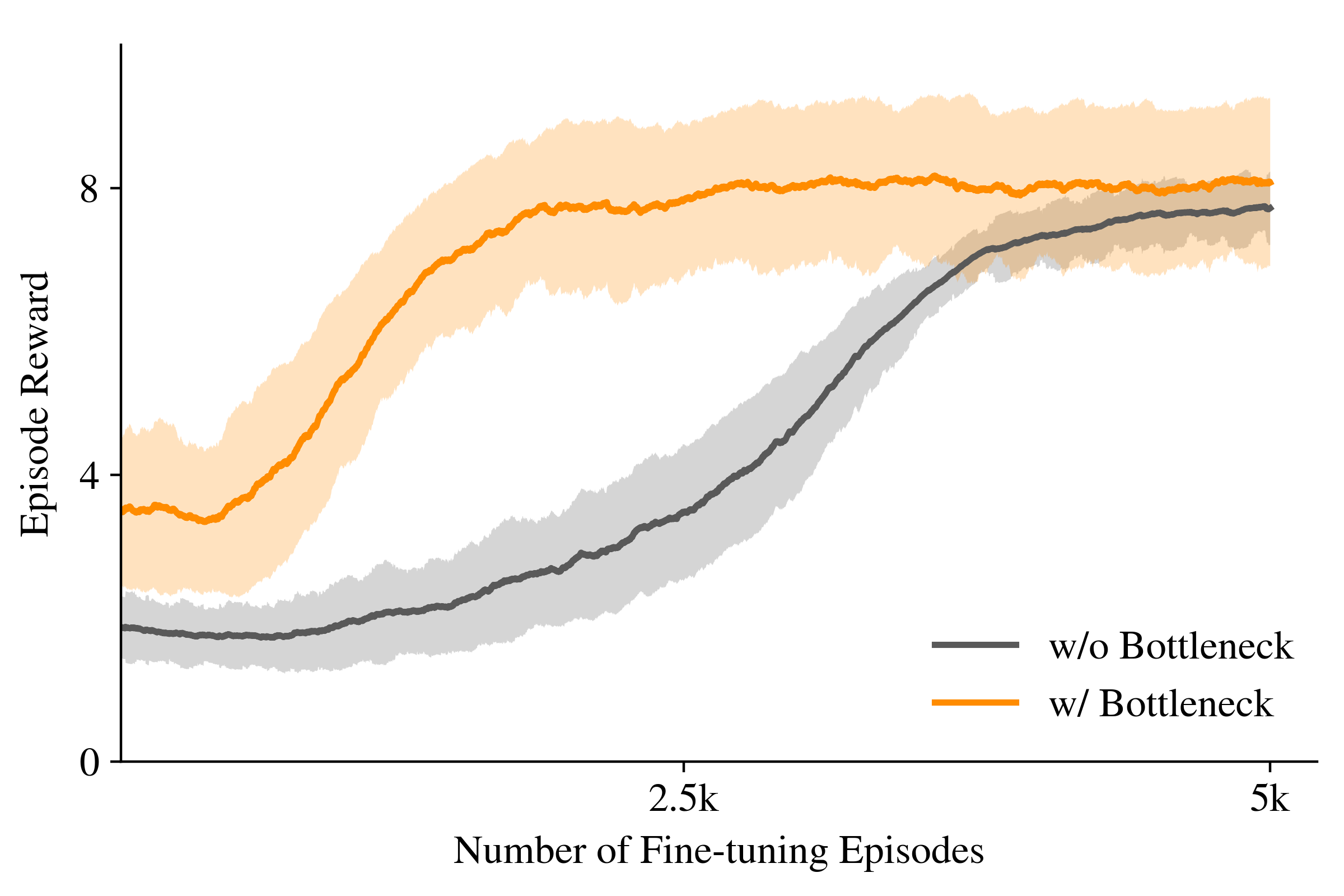}
  \caption{{Training a policy on a dataset of intersections and highways and transferring it to roundabouts. Using a latent representation speeds up the transfer learning.}}
  \label{fig:transfer_learning}
\end{figure}
\begin{figure*}[t]
  \centering
  \includegraphics[width=.9\textwidth]{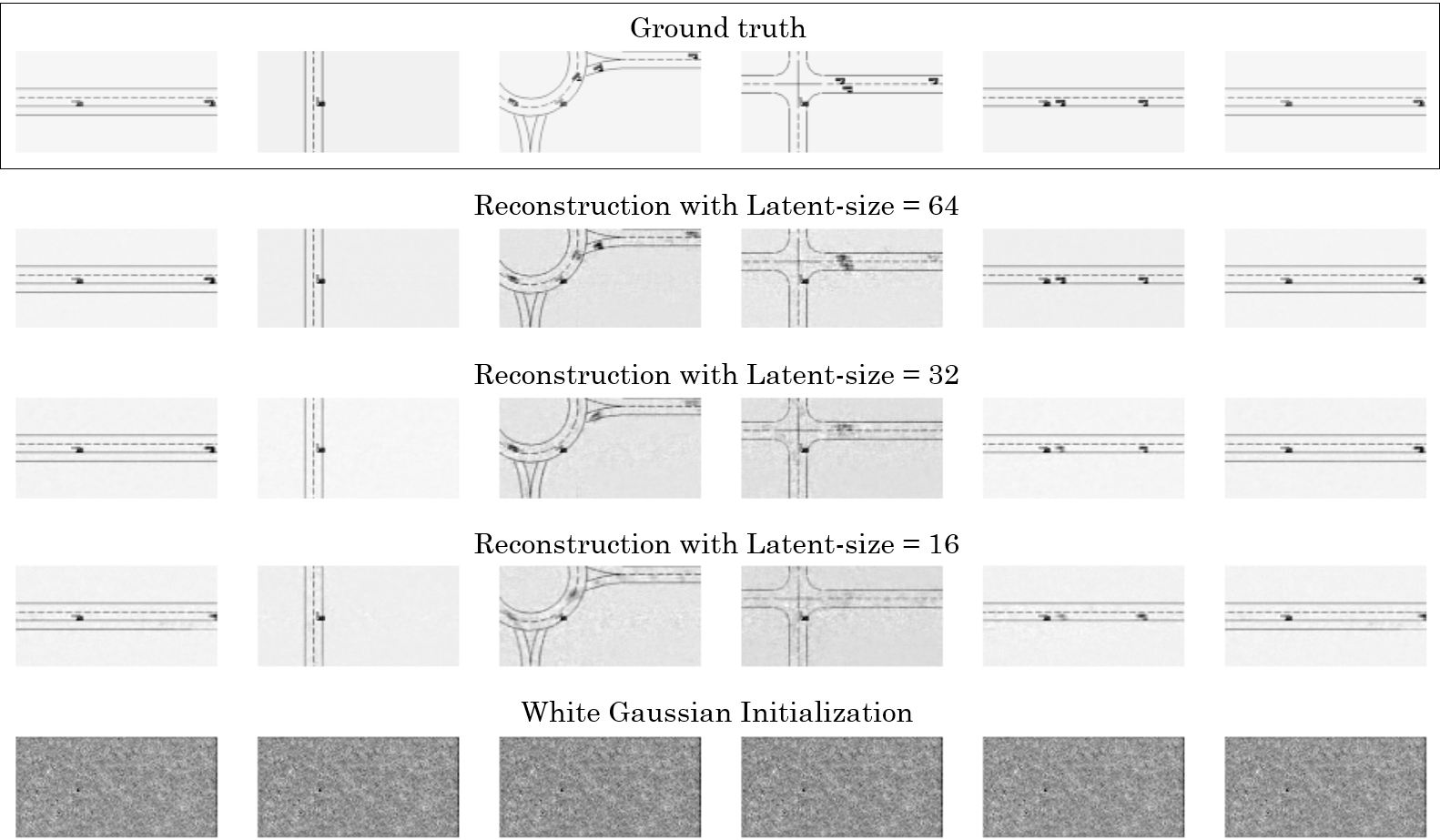}
  \caption{{Choosing the size of the latent space to be too small forces an excessive level of compression that might result in losing crucial and necessary information about the scene. We compare the reconstruction capability of the bottleneck with different sizes of latent space.}}
  \label{fig:reconstruction_fullpage}
\end{figure*}

\subsection{Driving Simulator}
\label{sec:drivingsimulator}
For the purpose of our simulations and generating a training dataset for the training of the autoencoders as well as the Q-learning process, we use an abstract 2D driving simulator based on an OpenAI Gym environment developed by Leurent\etal~\cite{leurent2019approximate}. Our simulator is able to generate diverse driving scenarios for a given road topology and vehicle placement. As our goal is to learn generalizable policies rather than memorizing a sequence of actions by VFAN, the initial state of each simulation episode is randomized. Initial Frenet latitude of the vehicles are drawn from a uniform distribution uniformly randomized and the initial Frenet longitude $l_{\mathcal{M}}(t_0)$ and Frenet longitudinal velocity $v_{\mathcal{M}}(t_0)$ of simulated vehicles are drawn from a clipped-Gaussian distribution as follows, 
\begin{equation}
\label{equ:clippedgaussian}
\widetilde{\mathcal{N}}(x)=\mathcal{N}(x;\mu,\sigma) \big[\mathds{1}(x-\mu+\delta) + \mathds{1} (x-\mu+\delta \big]
\end{equation}
where $\mathcal{N}(x;\mu,\sigma)$ denotes a Gaussian distribution and $\mathds{1}$ is the Heaviside step function. As we discussed before, the meta-actions of each vehicle are rendered into low-level steering and acceleration signals via employing a closed-loop PID controller. A Kinematic Bicycle Model then determines the vehicles' yaw rate based on the steering angle and other parameters.

Our simulations include both autonomous and human-driven vehicles to create a realistic mixed-autonomy scenario. We employ two widely-used human driver models proposed by Treiber\etal and Kesting\etal~\cite{treiber2000congested, kesting2007general}. Lateral actions of HVs and their decision to make a lane change, follow the Minimizing Overall Braking Induced by Lane changes (MOBIL) strategy~\cite{kesting2007general}. MOBIL model allows a lane change only if the resulting acceleration in the following vehicle meets the safety criterion. The longitudinal acceleration of HVs follows the Intelligent Driver Model (IDM)~\cite{treiber2000congested}. 

\subsection{Results}
\subsubsection{Learning Similarity-maximizing Bottleneck Network}
As we elaborated in Section~\ref{sec:solution}, we rely on the autoencoder architecture introduced in Section~\ref{sec:denoising_ae} to impose an information bottleneck and learn a latent representation that later will help to learn generalizable driving policies. Figure~\ref{fig:ae_loss} shows the training and validation loss for 100 epochs over a large shuffled dataset that includes observations from various driving scenarios, e.g., roundabout, intersection, highway merging and highway exiting. As a reminder, the input to this bottleneck network is a noisy stacked VelocityMap and we are interested in the latent space between the encoder and the decoder. Figure~\ref{fig:latent_rep} demonstrates examples of the learned latent representation on two contrasting road topologies, i.e., intersection and roundabout. The weights of the encoder and decoder networks are initialized with random weights and it can be observed that before training, the latent representation is simply an effect of the network initialization. However, after training for 100 epochs over the whole shuffled dataset, the latent representation generated from the two road topologies diverge. As a reminder, the aim of training this latent space is to maximize the similarity between two contrasting driving scenes.

It is clear that the size of the latent space dictates the level of information compression in our encoder-decoder architecture. We aim to impose a bottleneck on the information flow in order to distill the state observation and extract the most crucial and abstract representation that enables the resulting policy to generalize to unseen and novel environments. However, it is important to ensure the fidelity and trustworthiness of this abstract representation. Choosing the size of the latent space to be too small forces an excessive level of compression that might result in losing crucial and necessary information about the scene. In Figure~\ref{fig:reconstruction_fullpage}, we compare the reconstruction capability of the bottleneck with different sizes of latent space. This experiment is done to empirically choose the appropriate size for the latent space that can still preserve the important information carried by the state observation. Comparing the ground truth flatten observation (top row) with rows 2-4, one can observe a destructive and severe information loss in the smaller latent space (16 nodes). Particularly, some of the vehicles in both the intersection and roundabout scenarios are faded or even totally lost in the reconstructed image. On the other hand, the larger latent space (64 nodes) preserves all the vehicles, despite inaccurate edges and some of the vehicles being faded. Hereafter, we set the size of the latent size to 64 and train the autoencoder over the full dataset.

\subsubsection{Generalization via Learned Latent Representations}
This section contains the main results of this paper. Our main purpose in this work was to improve the generalization capabilities of our proposed restricted bottleneck architecture. We compare the performance of the learned policies in terms of the percentage of driving scenarios that experience a crash. The chosen baseline (w/o bottleneck) uses the exact same VFAN that is directly fed with stacked VelocityMaps. It is important to note that after learning the bottleneck network, the weights of the encoder are frozen and only the VFAN weights are trained during Q-learning. Hence, the comparison between the two methods with and without bottleneck is fair as they have identical capacity in VFAN and only differ in the input representation. 

\begin{figure}[t]
  \centering

  \includegraphics[width=0.49\textwidth]{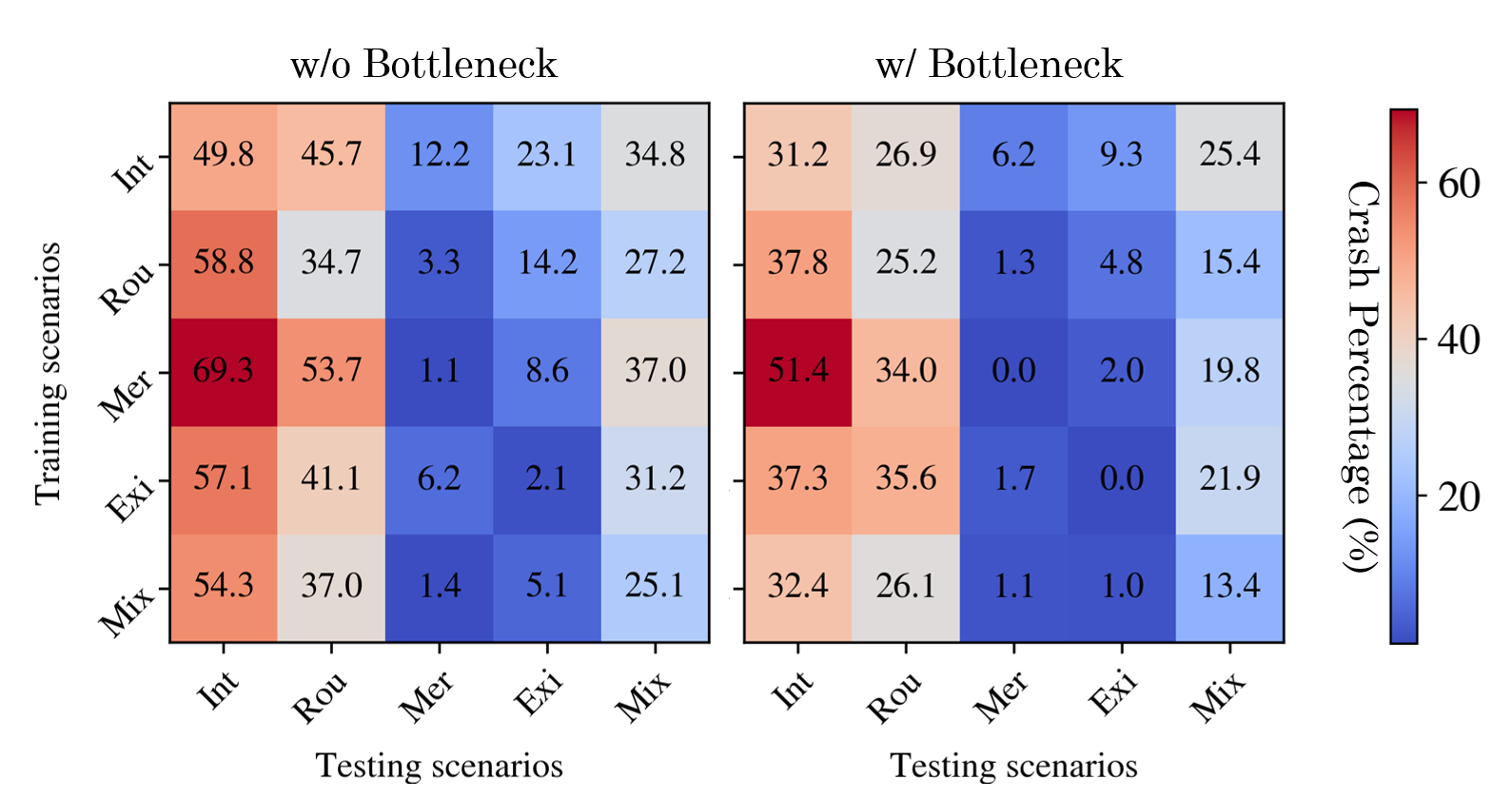}
  \caption{{Performance comparison between baseline and bottleneck networks across different transfer learning modes.}}
  \label{fig:perf_comparison}
\end{figure}

Figure~\ref{fig:effect_bottleneck} shows the performance enhancement resulted from the restricted latent representations in all four driving scenarios. More importantly, when the learned policy is tested on a mixed dataset that contains instances from roundabout, intersection, highway exiting, and highway merging, it leads to 47\% less crashes and failed episodes. Additionally, some road topologies are more challenging in nature, specifically, roundabout and intersection seem to experience a significantly higher number of crashes. Even when the policy is trained and tested on the same scenario, e.g., intersection, performance greatly improves by using our learned latent representation. This observation reveals the fact that not only our latent representation improves generalization across different driving scenarios, but it also helps learning safer and more efficient driving policies due to the distilled state representation as compared to the raw VelocityMaps. Figure~\ref{fig:perf_comparison} further illustrates the transfer learning capabilities of the learned driving in the form of the confusion matrices. It is evident that across all different combinations of test and training episodes, using the bottleneck network helps to lessen the number of collisions.

\subsubsection{Transfer Learning \& Domain Adaptation}

\begin{figure}[t]
  \centering

  \includegraphics[width=0.5\textwidth]{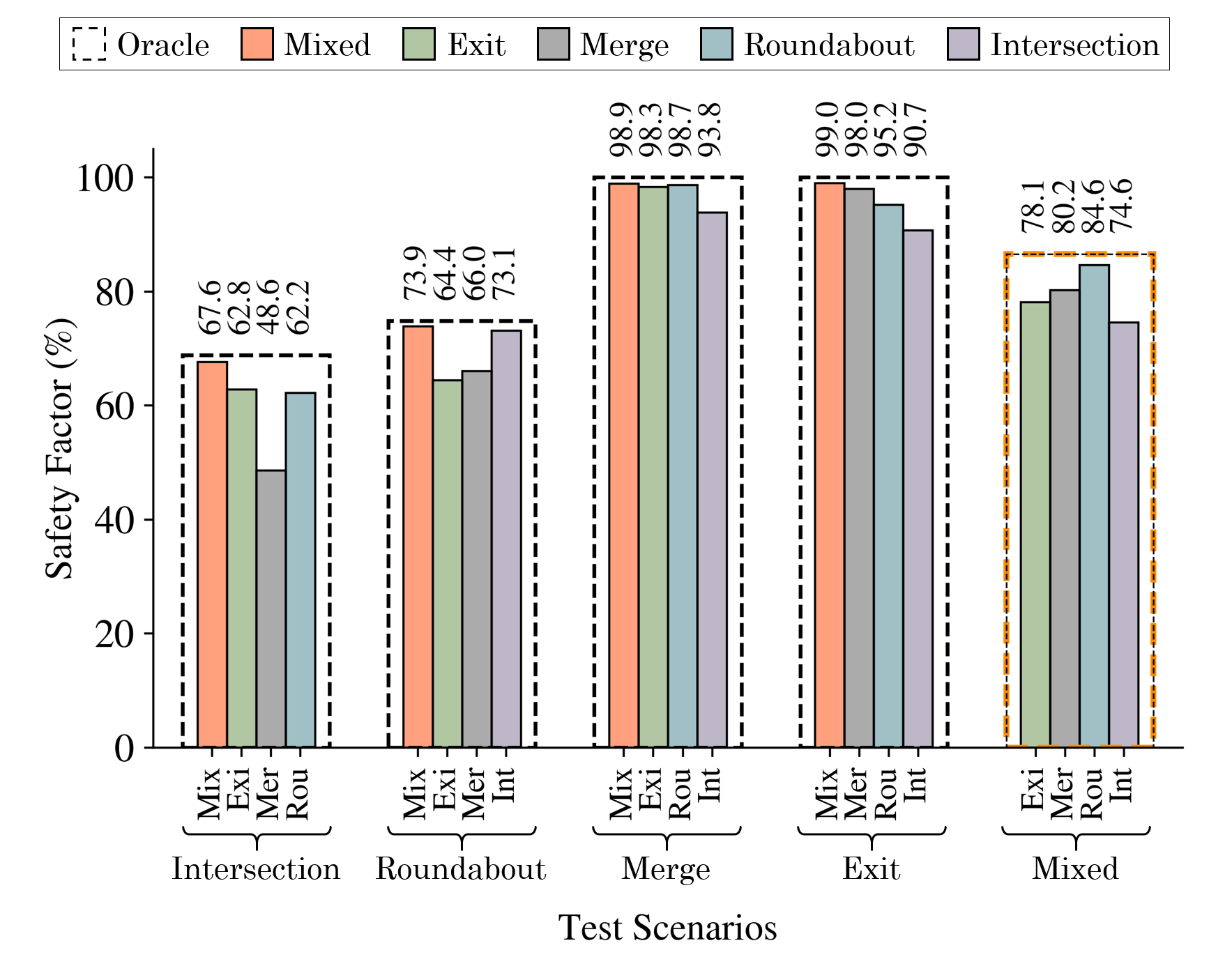}
  \caption{{Domain Adaptation.}}
  \label{fig:domain_adaptation}
\end{figure}

As an additional important investigation, we gauge the transfer learning capabilities of the learned policy with and without using the bottleneck network. We first train a policy over a shuffled dataset that contains highway merging and exiting, and intersection and then fine-tune the policy on the target scenario, i.e., roundabout. This experiment is performed with and without the bottleneck network. Figure~\ref{fig:transfer_learning} interestingly demonstrates that the policy that uses the latent representations as its input during the training, converges much faster in transfer learning as compared to the baseline.

Figure~\ref{fig:domain_adaptation} provides further intuition on transfer learning capabilities of the policy that is learned using the latent representations. When studying domain adaptation and generalization capabilities of driving policies, the best-expected performance happens when the policy is trained and tested in the same driving scenario, e.g., trained using randomized episodes of an intersection and tested on an intersection, we refer to this setup as the oracle and compare all other scenarios with the oracle. As shown in Figure~\ref{fig:domain_adaptation}, through training with the bottleneck on a mixed dataset that contains all scenarios, we can get very close to the performance of the oracle.
%
%
\section{Concluding Remarks}

\smallskip
\noindent \textbf{Summary. }
Human drivers perform fairly well in adapting to new driving scenarios and novel road topologies. We hypothesize that this capability results from constituting abstract representations of the driving scene and the environment by the driver in such a way that it grasps the most crucial information, such as free spaces, drivable areas, and distance from the center line of the road. In the case of autonomous vehicles, a driving policy that is learned in a given set of driving setups and road topologies usually poorly generalizes to novel scenarios. To address this shortcoming, we seek a latent representation that can maximize the similarity between various driving scenarios and hence boost the generalization ability of a learned driving policy. We choose an encoder-decoder structure to impose a bottleneck on the information flow and force the network to only pass the most crucial and abstract information. This latent representation is then used in conjunction with a 3D convolutional Q-learning module to learn driving policies. Our experiments reveal that through using this bottleneck structure, the number of collisions can be reduced by 50\%. Additionally, using our method improves the transfer learning capability of the learned policy and accelerates the convergence.

\smallskip
\noindent \textbf{Limitations and Future Work. }
In this work, we generated a large shuffled dataset of driving episodes from various road topologies, e.g., roundabouts, intersections, and highways, using our driving simulator. However, a more sophisticated study can be done on longer episodes of driving that are performed on randomly generated roads with random vehicle placements that include diverse segments, including the ones mentioned earlier. Additionally, more work must be done to interpret the learned latent representations and help us to build intuition on the information that is embedded in the latent space. Further exploration of these topics is planned for our future work.

%
\ifCLASSOPTIONcaptionsoff
  \newpage
\fi

\bibliographystyle{IEEEtran}
\bibliography{IEEEbibs}

\begin{thebibliography}{10}
\providecommand{\url}[1]{#1}
\csname url@samestyle\endcsname
\providecommand{\newblock}{\relax}
\providecommand{\bibinfo}[2]{#2}
\providecommand{\BIBentrySTDinterwordspacing}{\spaceskip=0pt\relax}
\providecommand{\BIBentryALTinterwordstretchfactor}{4}
\providecommand{\BIBentryALTinterwordspacing}{\spaceskip=\fontdimen2\font plus
\BIBentryALTinterwordstretchfactor\fontdimen3\font minus
  \fontdimen4\font\relax}
\providecommand{\BIBforeignlanguage}[2]{{%
\expandafter\ifx\csname l@#1\endcsname\relax
\typeout{** WARNING: IEEEtran.bst: No hyphenation pattern has been}%
\typeout{** loaded for the language `#1'. Using the pattern for}%
\typeout{** the default language instead.}%
\else
\language=\csname l@#1\endcsname
\fi
#2}}
\providecommand{\BIBdecl}{\relax}
\BIBdecl

\bibitem{toghi2021social}
B.~Toghi, R.~Valiente, D.~Sadigh, R.~Pedarsani, and Y.~P. Fallah, ``Social
  coordination and altruism in autonomous driving,'' \emph{arXiv preprint
  arXiv:2107.00200}, 2021.

\bibitem{toghi2018multiple}
B.~Toghi, M.~Saifuddin, H.~N. Mahjoub, M.~Mughal, Y.~P. Fallah, J.~Rao, and
  S.~Das, ``Multiple access in cellular v2x: Performance analysis in highly
  congested vehicular networks,'' in \emph{2018 IEEE Vehicular Networking
  Conference (VNC)}.\hskip 1em plus 0.5em minus 0.4em\relax IEEE, 2018, pp.
  1--8.

\bibitem{toghi2019analysis}
B.~Toghi, M.~Saifuddin, Y.~P. Fallah, and M.~Mughal, ``Analysis of distributed
  congestion control in cellular vehicle-to-everything networks,'' in
  \emph{2019 IEEE 90th Vehicular Technology Conference (VTC2019-Fall)}.\hskip
  1em plus 0.5em minus 0.4em\relax IEEE, 2019, pp. 1--7.

\bibitem{toghi2019spatio}
B.~Toghi, M.~Saifuddin, M.~Mughal, and Y.~P. Fallah, ``Spatio-temporal dynamics
  of cellular v2x communication in dense vehicular networks,'' in \emph{2019
  IEEE 2nd Connected and Automated Vehicles Symposium (CAVS)}.\hskip 1em plus
  0.5em minus 0.4em\relax IEEE, 2019, pp. 1--5.

\bibitem{toghi2020maneuver}
B.~Toghi, D.~Grover, M.~Razzaghpour, R.~Jain, R.~Valiente, M.~Zaman, G.~Shah,
  and Y.~P. Fallah, ``A maneuver-based urban driving dataset and model for
  cooperative vehicle applications,'' 2020.

\bibitem{8690570}
H.~N. {Mahjoub}, B.~{Toghi}, and Y.~P. {Fallah}, ``A stochastic hybrid
  framework for driver behavior modeling based on hierarchical dirichlet
  process,'' in \emph{2018 IEEE 88th Vehicular Technology Conference
  (VTC-Fall)}, 2018, pp. 1--5.

\bibitem{saifuddin2020performance}
M.~Saifuddin, M.~Zaman, B.~Toghi, Y.~P. Fallah, and J.~Rao, ``Performance
  analysis of cellular-v2x with adaptive \& selective power control,'' in
  \emph{2020 IEEE 3rd Connected and Automated Vehicles Symposium (CAVS)}.\hskip
  1em plus 0.5em minus 0.4em\relax IEEE, 2020, pp. 1--7.

\bibitem{mahjoub2019v2x}
H.~N. Mahjoub, B.~Toghi, S.~O. Gani, and Y.~P. Fallah, ``V2x system
  architecture utilizing hybrid gaussian process-based model structures,'' in
  \emph{2019 IEEE International Systems Conference (SysCon)}.\hskip 1em plus
  0.5em minus 0.4em\relax IEEE, 2019, pp. 1--7.

\bibitem{shah2019real}
G.~Shah, R.~Valiente, N.~Gupta, S.~O. Gani, B.~Toghi, Y.~P. Fallah, and S.~D.
  Gupta, ``Real-time hardware-in-the-loop emulation framework for dsrc-based
  connected vehicle applications,'' in \emph{2019 IEEE 2nd Connected and
  Automated Vehicles Symposium (CAVS)}.\hskip 1em plus 0.5em minus 0.4em\relax
  IEEE, 2019, pp. 1--6.

\bibitem{mahjoub2018driver}
H.~N. Mahjoub, B.~Toghi, and Y.~P. Fallah, ``A driver behavior modeling
  structure based on non-parametric bayesian stochastic hybrid architecture,''
  in \emph{2018 IEEE 88th Vehicular Technology Conference (VTC-Fall)}.\hskip
  1em plus 0.5em minus 0.4em\relax IEEE, 2018, pp. 1--5.

\bibitem{valiente2019controlling}
R.~Valiente, M.~Zaman, S.~Ozer, and Y.~P. Fallah, ``Controlling steering angle
  for cooperative self-driving vehicles utilizing cnn and lstm-based deep
  networks,'' in \emph{2019 IEEE Intelligent Vehicles Symposium (IV)}.\hskip
  1em plus 0.5em minus 0.4em\relax IEEE, 2019, pp. 2423--2428.

\bibitem{valiente2020connected}
R.~Valiente, M.~Zaman, Y.~P. Fallah, and S.~Ozer, ``Connected and autonomous
  vehicles in the deep learning era: A case study on computer-guided
  steering,'' in \emph{Handbook Of Pattern Recognition And Computer
  Vision}.\hskip 1em plus 0.5em minus 0.4em\relax World Scientific, 2020, pp.
  365--384.

\bibitem{toghi2021cooperative}
B.~Toghi, R.~Valiente, D.~Sadigh, R.~Pedarsani, and Y.~P. Fallah, ``Cooperative
  autonomous vehicles that sympathize with human drivers,'' \emph{arXiv
  preprint arXiv:2107.00898}, 2021.

\bibitem{toghi2021altruistic}
------, ``Altruistic maneuver planning for cooperative autonomous vehicles
  using multi-agent advantage actor-critic,'' \emph{arXiv preprint
  arXiv:2107.05664}, 2021.

\bibitem{azimi2013reliable}
S.~Azimi, G.~Bhatia, R.~Rajkumar, and P.~Mudalige, ``Reliable intersection
  protocols using vehicular networks,'' in \emph{Proceedings of the ACM/IEEE
  4th International Conference on Cyber-Physical Systems}, 2013, pp. 1--10.

\bibitem{azimi2015ballroom}
R.~Azimi, G.~Bhatia, R.~Rajkumar, and P.~Mudalige, ``Ballroom intersection
  protocol: Synchronous autonomous driving at intersections,'' in \emph{2015
  IEEE 21st International Conference on Embedded and Real-Time Computing
  Systems and Applications}.\hskip 1em plus 0.5em minus 0.4em\relax IEEE, 2015,
  pp. 167--175.

\bibitem{wang2017formulation}
P.~Wang and C.-Y. Chan, ``Formulation of deep reinforcement learning
  architecture toward autonomous driving for on-ramp merge,'' in \emph{2017
  IEEE 20th International Conference on Intelligent Transportation Systems
  (ITSC)}.\hskip 1em plus 0.5em minus 0.4em\relax IEEE, 2017, pp. 1--6.

\bibitem{dong2017intention}
C.~Dong, J.~M. Dolan, and B.~Litkouhi, ``Intention estimation for ramp merging
  control in autonomous driving,'' in \emph{2017 IEEE intelligent vehicles
  symposium (IV)}.\hskip 1em plus 0.5em minus 0.4em\relax IEEE, 2017, pp.
  1584--1589.

\bibitem{hoermann2018dynamic}
S.~Hoermann, M.~Bach, and K.~Dietmayer, ``Dynamic occupancy grid prediction for
  urban autonomous driving: A deep learning approach with fully automatic
  labeling,'' in \emph{2018 IEEE International Conference on Robotics and
  Automation (ICRA)}.\hskip 1em plus 0.5em minus 0.4em\relax IEEE, 2018, pp.
  2056--2063.

\bibitem{chen2019model}
J.~Chen, B.~Yuan, and M.~Tomizuka, ``Model-free deep reinforcement learning for
  urban autonomous driving,'' in \emph{2019 IEEE intelligent transportation
  systems conference (ITSC)}.\hskip 1em plus 0.5em minus 0.4em\relax IEEE,
  2019, pp. 2765--2771.

\bibitem{chen2021interpretable}
J.~Chen, S.~E. Li, and M.~Tomizuka, ``Interpretable end-to-end urban autonomous
  driving with latent deep reinforcement learning,'' \emph{IEEE Transactions on
  Intelligent Transportation Systems}, 2021.

\bibitem{zhan2016non}
W.~Zhan, C.~Liu, C.-Y. Chan, and M.~Tomizuka, ``A non-conservatively defensive
  strategy for urban autonomous driving,'' in \emph{2016 IEEE 19th
  International Conference on Intelligent Transportation Systems (ITSC)}.\hskip
  1em plus 0.5em minus 0.4em\relax IEEE, 2016, pp. 459--464.

\bibitem{huang2019motion}
Y.~Huang, H.~Ding, Y.~Zhang, H.~Wang, D.~Cao, N.~Xu, and C.~Hu, ``A motion
  planning and tracking framework for autonomous vehicles based on artificial
  potential field elaborated resistance network approach,'' \emph{IEEE
  Transactions on Industrial Electronics}, vol.~67, no.~2, pp. 1376--1386,
  2019.

\bibitem{chen2017constrained}
J.~Chen, W.~Zhan, and M.~Tomizuka, ``Constrained iterative lqr for on-road
  autonomous driving motion planning,'' in \emph{2017 IEEE 20th International
  Conference on Intelligent Transportation Systems (ITSC)}.\hskip 1em plus
  0.5em minus 0.4em\relax IEEE, 2017, pp. 1--7.

\bibitem{chen2019autonomous}
------, ``Autonomous driving motion planning with constrained iterative lqr,''
  \emph{IEEE Transactions on Intelligent Vehicles}, vol.~4, no.~2, pp.
  244--254, 2019.

\bibitem{tschannen2018recent}
M.~Tschannen, O.~Bachem, and M.~Lucic, ``Recent advances in autoencoder-based
  representation learning,'' \emph{arXiv preprint arXiv:1812.05069}, 2018.

\bibitem{plebe2020road}
A.~Plebe and M.~Da~Lio, ``On the road with 16 neurons: Towards interpretable
  and manipulable latent representations for visual predictions in driving
  scenarios,'' \emph{IEEE Access}, vol.~8, pp. 179\,716--179\,734, 2020.

\bibitem{li2017variation}
H.~Li, H.~Wang, Z.~Yang, and M.~Odagaki, ``Variation autoencoder based network
  representation learning for classification,'' in \emph{Proceedings of ACL
  2017, Student Research Workshop}, 2017, pp. 56--61.

\bibitem{chen2020end}
J.~Chen, Z.~Xu, and M.~Tomizuka, ``End-to-end autonomous driving perception
  with sequential latent representation learning,'' in \emph{2020 IEEE/RSJ
  International Conference on Intelligent Robots and Systems (IROS)}.\hskip 1em
  plus 0.5em minus 0.4em\relax IEEE, 2020, pp. 1999--2006.

\bibitem{kargar2020efficient}
E.~Kargar and V.~Kyrki, ``Efficient latent representations using multiple tasks
  for autonomous driving,'' \emph{arXiv preprint arXiv:2003.00695}, 2020.

\bibitem{ma2021reinforcement}
X.~Ma, J.~Li, M.~J. Kochenderfer, D.~Isele, and K.~Fujimura, ``Reinforcement
  learning for autonomous driving with latent state inference and
  spatial-temporal relationships,'' in \emph{2021 IEEE International Conference
  on Robotics and Automation (ICRA)}.\hskip 1em plus 0.5em minus 0.4em\relax
  IEEE, 2021, pp. 6064--6071.

\bibitem{xie2020learning}
A.~Xie, D.~Losey, R.~Tolsma, C.~Finn, and D.~Sadigh, ``Learning latent
  representations to influence multi-agent interaction,'' in \emph{Proceedings
  of the 4th Conference on Robot Learning (CoRL)}, November 2020.

\bibitem{losey2021learning}
D.~P. Losey, H.~J. Jeon, M.~Li, K.~Srinivasan, A.~Mandlekar, A.~Garg, J.~Bohg,
  and D.~Sadigh, ``Learning latent actions to control assistive robots,''
  \emph{Autonomous robots}, pp. 1--33, 2021.

\bibitem{karamcheti2021lila}
S.~Karamcheti, M.~Srivastava, P.~Liang, and D.~Sadigh, ``Lila:
  Language-informed latent actions,'' in \emph{5th Annual Conference on Robot
  Learning}, 2021.

\bibitem{karamcheti2021learning}
S.~Karamcheti, A.~J. Zhai, D.~P. Losey, and D.~Sadigh, ``Learning visually
  guided latent actions for assistive teleoperation,'' in \emph{Learning for
  Dynamics and Control}.\hskip 1em plus 0.5em minus 0.4em\relax PMLR, 2021, pp.
  1230--1241.

\bibitem{lee2020optimization}
D.~Lee, N.~He, P.~Kamalaruban, and V.~Cevher, ``Optimization for reinforcement
  learning: From a single agent to cooperative agents,'' \emph{IEEE Signal
  Processing Magazine}, vol.~37, no.~3, pp. 123--135, 2020.

\bibitem{mnih2013playing}
V.~Mnih, K.~Kavukcuoglu, D.~Silver, A.~Graves, I.~Antonoglou, D.~Wierstra, and
  M.~Riedmiller, ``Playing atari with deep reinforcement learning,''
  \emph{arXiv preprint arXiv:1312.5602}, 2013.

\bibitem{vincent2008extracting}
P.~Vincent, H.~Larochelle, Y.~Bengio, and P.-A. Manzagol, ``Extracting and
  composing robust features with denoising autoencoders,'' in \emph{Proceedings
  of the 25th international conference on Machine learning}, 2008, pp.
  1096--1103.

\bibitem{gehring2013extracting}
J.~Gehring, Y.~Miao, F.~Metze, and A.~Waibel, ``Extracting deep bottleneck
  features using stacked auto-encoders,'' in \emph{2013 IEEE international
  conference on acoustics, speech and signal processing}.\hskip 1em plus 0.5em
  minus 0.4em\relax IEEE, 2013, pp. 3377--3381.

\bibitem{schwarting2019social}
W.~Schwarting, A.~Pierson, J.~Alonso-Mora, S.~Karaman, and D.~Rus, ``Social
  behavior for autonomous vehicles,'' \emph{Proceedings of the National Academy
  of Sciences}, vol. 116, no.~50, pp. 24\,972--24\,978, 2019.

\bibitem{gupta2017cognitive}
S.~Gupta, J.~Davidson, S.~Levine, R.~Sukthankar, and J.~Malik, ``Cognitive
  mapping and planning for visual navigation,'' in \emph{Proceedings of the
  IEEE Conference on Computer Vision and Pattern Recognition}, 2017, pp.
  2616--2625.

\bibitem{leurent2019approximate}
E.~Leurent, Y.~Blanco, D.~Efimov, and O.-A. Maillard, ``Approximate robust
  control of uncertain dynamical systems,'' \emph{arXiv preprint
  arXiv:1903.00220}, 2019.

\bibitem{treiber2000congested}
M.~Treiber, A.~Hennecke, and D.~Helbing, ``Congested traffic states in
  empirical observations and microscopic simulations,'' \emph{Physical review
  E}, vol.~62, no.~2, p. 1805, 2000.

\bibitem{kesting2007general}
A.~Kesting, M.~Treiber, and D.~Helbing, ``General lane-changing model mobil for
  car-following models,'' \emph{Transportation Research Record}, vol. 1999,
  no.~1, pp. 86--94, 2007.

\end{thebibliography}

\vskip -2.5\baselineskip plus -1fil
\begin{IEEEbiography}[{\includegraphics[width=1in,height=1.25in,clip,keepaspectratio]{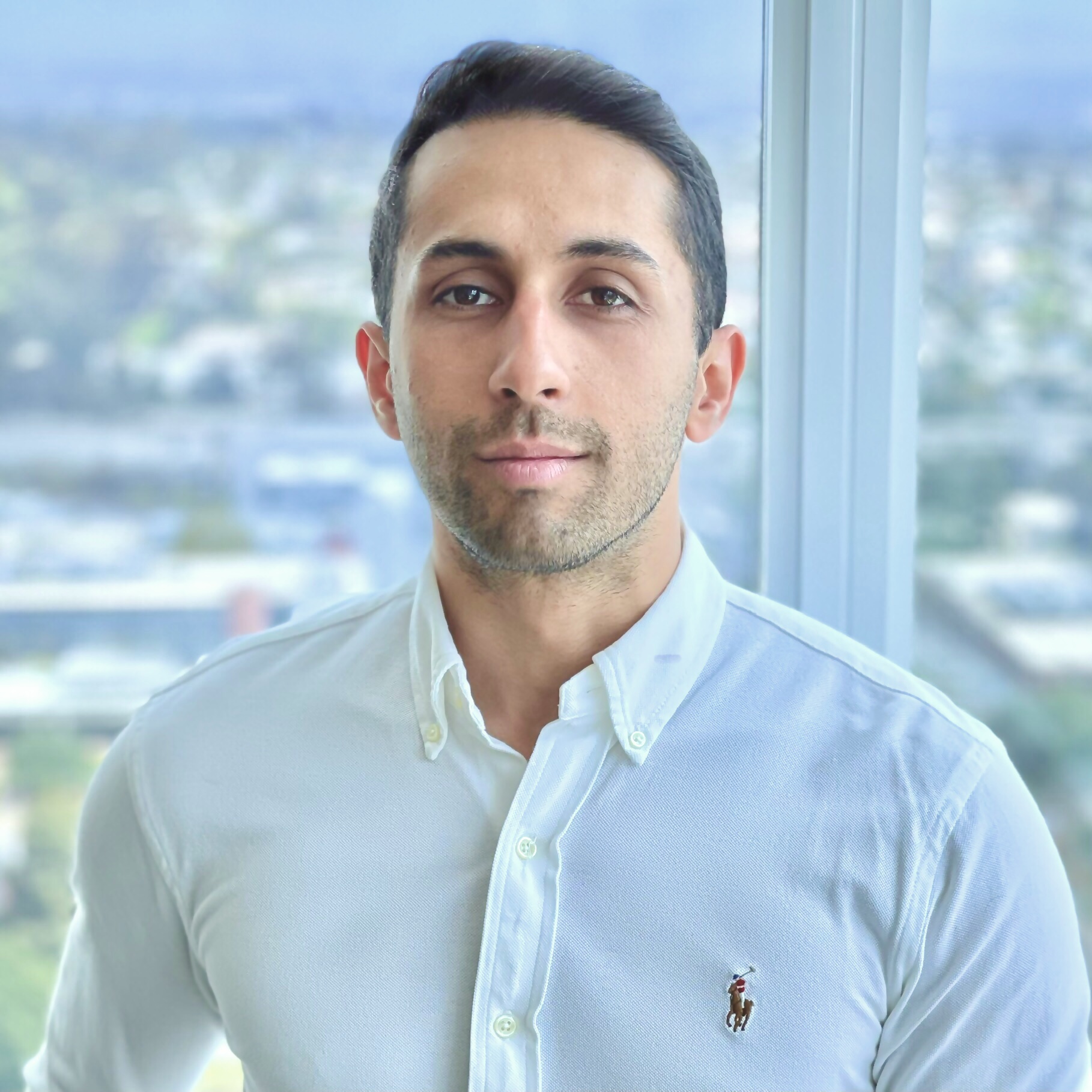}}]{Behrad Toghi}
is a Ph.D. candidate at the University of Central Florida. He received the B.Sc. degree in electrical engineering from Sharif University of Technology in 2016 and has worked as a research intern at Honda Research Institute, Mercedes-Benz R\&D North America and Ford Motor Company R\&D between 2018 and 2021. His work is mainly in the domains of prediction \& behavior planning for autonomous driving.
\end{IEEEbiography}
\vskip -2.5\baselineskip plus -1fil
\begin{IEEEbiography}[{\includegraphics[width=1in,height=1.25in,clip,keepaspectratio]{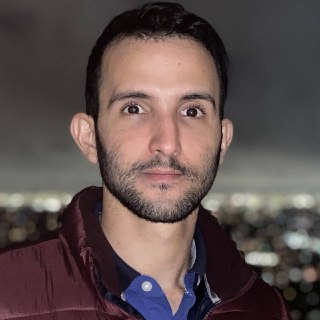}}]{Rodolfo Valiente}
is a Ph.D. candidate in Computer Engineering at the University of Central Florida. His research interests include connected autonomous vehicles (CAVs), reinforcement learning, computer vision, and deep learning with a focus on the autonomous driving problem. He received a M.Sc. degree from the University of Sao Paulo and his B.Sc. degree from the Technological University Jose Antonio Echeverria.
\end{IEEEbiography}
\vskip -2.5\baselineskip plus -1fil
\begin{IEEEbiography}[{\includegraphics[width=1in,height=1.25in,clip,keepaspectratio]{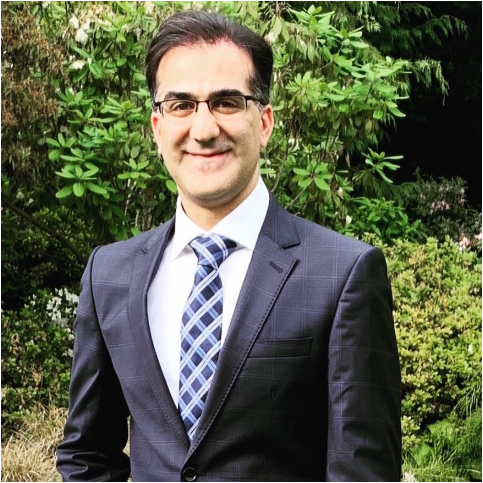}}]{Ramtin Pedarsani}
is an Assistant Professor in the ECE Department at the University of California, Santa Barbara. He received the B.Sc. degree in electrical engineering from the University of Tehran in 2009, the M.Sc. degree in communication systems from the Swiss Federal Institute of Technology (EPFL) in 2011, and his Ph.D. from the University of California, Berkeley, in 2015. His research interests include networks, game theory, machine learning, and transportation systems.
\end{IEEEbiography}
\vskip -2.5\baselineskip plus -1fil
\begin{IEEEbiography}[{\includegraphics[width=1in,height=1.25in,clip,keepaspectratio]{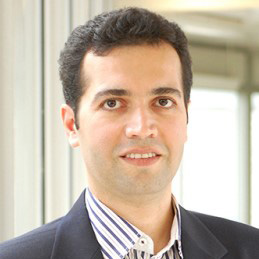}}]{Yaser P. Fallah} is an Associate Professor in the ECE Department at the University of Central Florida. He received the Ph.D. degree from the University of British Columbia, Vancouver, BC, Canada, in 2007. From 2008 to 2011, he was a Research Scientist with the Institute of Transportation Studies,
University of California Berkeley, Berkeley, CA, USA. His research, sponsored by industry, USDoT, and NSF, is focused on intelligent transportation systems and automated and networked vehicle safety systems.
\end{IEEEbiography}

\end{document}